\documentclass[conference,a4paper]{IEEEtran}
\ifCLASSINFOpdf
\else
\fi
\usepackage{graphicx}
\usepackage{amsmath}
\usepackage{amssymb}
\usepackage{algorithm}
\usepackage{algorithmic}
\usepackage{sidecap}
\usepackage{cite}

\begin{document}

%
\title{Low-rank SIFT: \\An Affine Invariant Feature For Place Recognition}


\author{\IEEEauthorblockN{Harry Yang\IEEEauthorrefmark{1},
Shengnan Cai\IEEEauthorrefmark{1},
Jingdong Wang\IEEEauthorrefmark{2} and
Long Quan\IEEEauthorrefmark{1} }
\IEEEauthorblockA{\IEEEauthorrefmark{1}Computer Science and Engineering Department\\
Hong Kong University of Science and Technology,
Hong Kong, China\\ Email: harryyang@ust.hk; scaiad@ust.hk; quan@cse.ust.hk}
\IEEEauthorblockA{\IEEEauthorrefmark{2}Microsoft Research Asia, No.5 Danling Street, Haidian District, Beijing 100080, China\\
Email: jingdw@microsoft.com}
}

\maketitle

\begin{abstract}
In this paper, we present a novel affine-invariant feature based on SIFT, leveraging
the regular appearance of man-made objects. The feature achieves full affine
invariance without needing to simulate over affine parameter space. Low-rank SIFT,
as we name the feature, is based on our observation that local tilt, which are caused by changes of camera axis orientation, could be normalized by converting local
patches to standard low-rank forms. Rotation, translation and scaling invariance could be achieved
in ways similar to SIFT. As an extension of SIFT, our method seeks to add prior to solve the
ill-posed affine parameter estimation problem and normalizes them directly, and is applicable to objects with regular structures. Furthermore, owing to recent breakthrough in convex
optimization, such parameter could be computed efficiently. We will demonstrate
its effectiveness in place recognition as our major application. As extra
contributions, we also describe our pipeline of constructing geotagged building database
from the ground up, as well as an efficient scheme for automatic feature selection.

\end{abstract}

\IEEEpeerreviewmaketitle

\section{Introduction}
Constructing features of
various invariance is the building block of image matching, which in turn is
fundamental to many computer vision and image processing tasks, such as camera
calibration, image retrieval, and place recognition. The
enthusiasm of seeking all kinds of invariance largely arises due to the complexity
of camera model, as different cameras and viewpoints lead to different
deformations for the same object (panorama image is an
example). Current image matching methods usually consist of three stages.
First, “interest points” are detected at descriptive locations, such
as corners or blobs. Next, an invariant descriptor is associated to region
around each interest point. This step is crucial as it is where the main challenge
lies to eliminate irrelevance and find the intrinsic local representation. Finally,
correspondence is established using descriptor vectors, and similar
images are retrieved. In this scenario, images of same object under
different transforms should have as similar feature representations as
possible.

\begin{figure}[t]
\begin{center}
\includegraphics[width=1\linewidth]{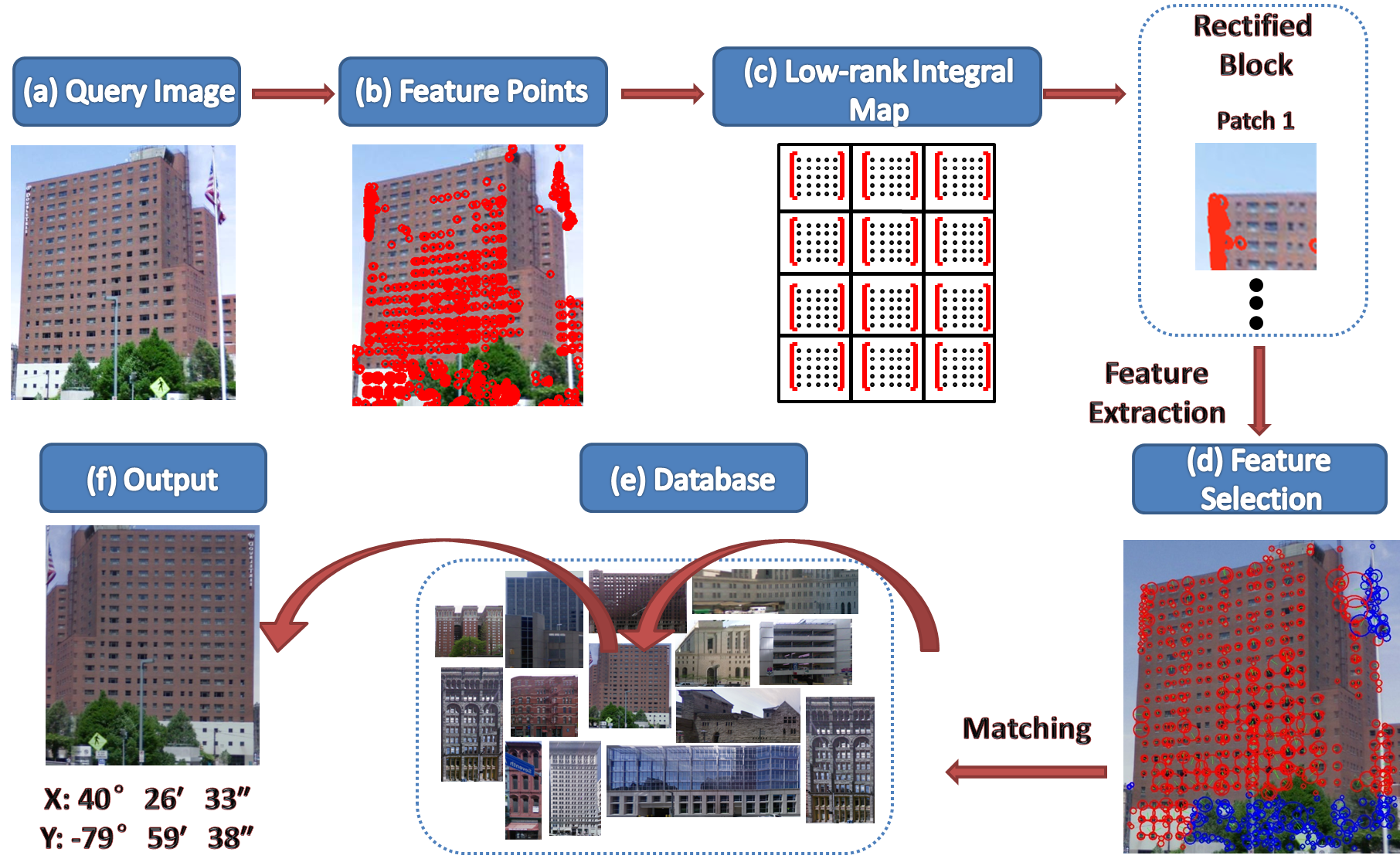}
\caption{Place recognition pipeline using low-rank SIFT: (a) Query image is fed into the pipeline. (b) Feature points are detected with Harris-Corner detector. (c) Low-rank transform is computed at block-level. (d) Features are selected based on the rank of patches, where features of fa\c{c}ade are colored in red and others in blue. SIFT descriptors are also computed accordingly. (e) Our place recognition database. (f) Recognition result returned with its geolocatoin. }
\label{flowchart}
\end{center}
\vspace{-0.4cm}
\end{figure}

Given the three-stage procedure there are two challenging questions to be answered. The first one, as we call it feature selection, is related to how to effectively select features of our interest from the entire feature set. Many investigations are done by exploiting the intrinsic property of features. Li et al.~\cite{li06} does filtering by assigning a posterior probability to each feature. A similar approach is~\cite{vidal2003}, which tries to detect informative fragments from image and uses them as features directly .~\cite{turcot09} selects features appearing repetitively in multiple images, while abandon those occasionally occur. All these methods lead to compressed and more representative feature set. Nevertheless,~\cite{SBS07} takes a different approach by using the feature information to guide the building of the vocabulary trees, and let the set of features remain the same. 

A more difficult question would be using what feature to achieve invariance, and recent years designing robust features have drawn huge attention. All the features proposed could be categorized based on the degree of invariance they achieved and what methods they use. One of the most notable feature is SIFT~\cite{lowe99}, which achieves partial affine invariance by normalizing translation and rotation, and is fully scale invariant by simulating over the scale space. Moment-based Harris-Affine\cite{harris02} and Hessian-Affine\cite{shmid04} aim to resolve viewpoint ambiguity, both of which detect corners or blobs first and then iteratively estimate the transform parameters. There are also region-based features such as MSER\cite{mser02} and LLD~\cite{lld03}, which try to normalize the most robust image level sets and level lines to get the standard form. ASIFT\cite{asift09} as an extension of SIFT, achieves full affine invariance, using both normalization and simulation of all the six affine parameters.

Our work is inspired by a recent global feature, the Transform Invariant Low-Rank Texture (TILT)\cite{tilt10}, which seeks global
invariance by transforming image to its low-rank form. Likewise, we build feature that could capture local structure as well as
low-level hue and gradient information. Our feature, as we name it Low-rank SIFT, is an extension of SIFT with two 
significant properties. First, it achieves full affine invariance, comparing with Harris-Affine, Hessian-Affine, MSER, etc. Second, it \textit{normalizes} the affine parameters rather than \textit{simulating} over the parameter space, comparing with ASIFT.

The main contributions of our work could be summarized as:
\begin{itemize}
\item We propose a novel framework that achieves affine-invariant feature representation and feature selection. The feature is fully affine invariant 
when applying to objects with regular appearance. 
\item We define a novel concept, the Low-rank Integral Map, and show how it 
enables local low-rank transform efficient.
\item We describe our method of building a benchmark database for place 
recognition. The database contains facades of multiple styles.
\end{itemize}
The pipeline of computing Low-rank SIFT and recognizing place is illustrated in Fig.~\ref{flowchart}.

\section{Normalizing the Tilt: An Overview}
Morel's efforts of introducing ASIFT and other features under a unified framework proves that, with the affine camera model, local perspective effects can be modeled by affine transform\cite{asift09}:
\[
u(x,y)\rightarrow u(ax+by+e,cx+dy+f).
\]
The affine transform could be further decomposed into the product of translation, scale, rotation and tilt. 

In spite their formulations are different, there are two different ways to achieve invariance against these transforms: normalization and simulation. In normalization, a local patch is converted to its standard form. In the example of SIFT, translation is easily normalized by translating a patch to the origin point. Rotation could be normalized by computing a principle direction and rotate the patch until its principle direction coincides with a fixed axis. Notwithstanding above, tilt or skew is more difficult to normalize. ASIFT uses simulation to perform all possible deformations and does matching using the entire simulated image set. Admittedly, given inadequate information, normalizing the tilt parameter is ill-posed as there is no standard form for random patch. However, if the local patch contains regular structure, such as significant horizontal/vertical lines, we can convert the patch to rectified shape, where horizontal lines go horizontal and vertical lines go vertical. In this regard, we can normalize tilt directly based on the following observation:

\begin{figure}[t]
\begin{center}
\includegraphics[width=0.9\linewidth]{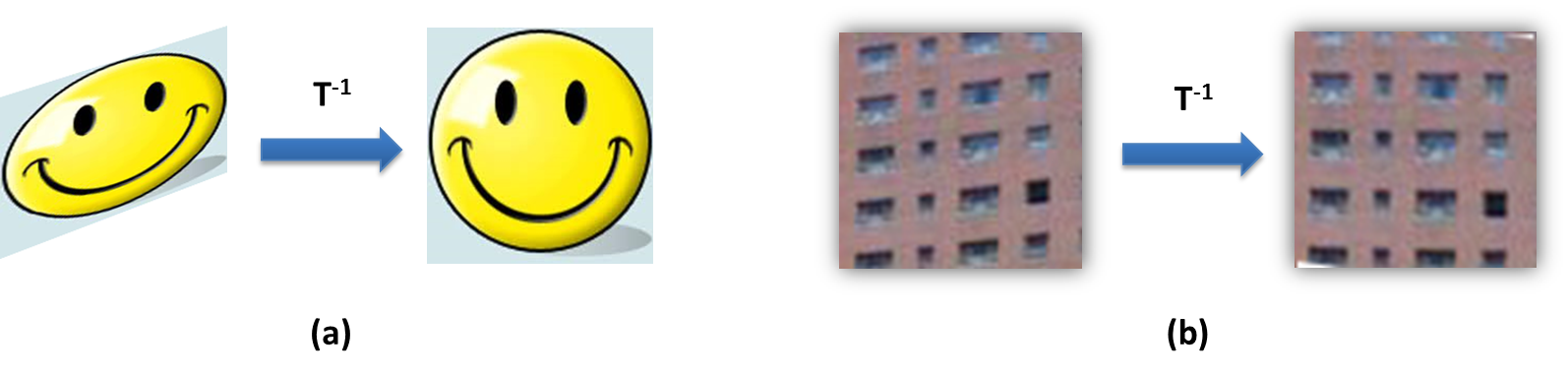}
\caption{(a). An object's perspective view and its rectified form. (b). A local patch could be transformed to low-rank shape with an affine transform.}
\label{demo}
\end{center}
\end{figure}
\textit{When a local patch is subject to regular appearance, it is possible to rectify the patch to its standard form with some affine transform.}

Fig.~\ref{demo} illustrates how an object and a local patch could be transformed to their standard rectified forms. In order to formulate it into a tangible problem, we observe that a rectified patch could usually be decomposed into a low-rank matrix with sparse noise. Similar to \cite{tilt10}, given a patch $I$, we would like to solve the following optimization problem:
\begin{equation}
  \min_{I_0,E,\tau}rank(I_0)+\lambda\|E\|_0 \text{  subject to } I\circ\tau=I_0 +E
\label{optimization}
\end{equation}
where $\lambda$ is a weight parameter, and the $l_0$ norm measures noise sparsity. $\tau$ is the transform we desire to compute. Our difference with the Transform Invariant Low-rank Textures (TILT) is that instead of seeking a global transform, we find low-rank structure locally in search of local invariance. Similar solver based on Augmented Lagrange Method (ALM) could be employed, but given the amount of feature points we have, this could be very time-consuming. We propose to use a novel Low-rank Integral Map to reduce the computational cost, a concept borrowed from SURF~\cite{surf06} where an integral image is computed as a preprocessing step. \\\\
\textbf{The Low-rank Prior.}
The success of Low-rank SIFT lies in the assumption of local regularity. In order to validate our assumption, we did a field test using our place recognition dataset. The dataset contains rectified building images of different styles. We randomly select a few images from several cities. For each image, we conduct an arbitrary prospective transformation to simulate random camera viewpoint. We then do local transform on surface of buildings, and rectifies the patch accordingly. The transformed patch is compared against the original one. As we expected, we see a big boost of similarity after performing low-rank transform on patches (Table~\ref{regularity}).
\begin{table}[ht]
\caption{Similarity of deformed patches with original ones before and after low-rank transform.}
\begin{center}
\resizebox{0.9\linewidth}{!}{
\begin{tabular}{ | c | c | c |}
\hline
City & Before Transform & After Transform \\ \hline
Paris & 0.68 & 0.96 \\ \hline
Pittsburgh & 0.65 & 0.94 \\ \hline
Hong Kong & 0.66 & 0.89 \\ \hline
\end{tabular}
}
\end{center}
\label{regularity}
\end{table}\\\\
\textbf{Fixing the Aspect Ratio.}
Tilt is caused by the change of the camera axis orientation and their relative scale, which involves two parameters. Low-rank transformation only normalizes the first one by forcing it to be mutually orthogonal but their relative scale remains unknown. For example, for a rectangle projected as a parallelogram, after a low-rank transform it is possible to be converted to a square. Fortunately in our experiment, we found this issue largely minor, and we can simply fix the aspect ratio, which has trivial impact on image matching results. This could be explained from two aspects. First, our interest point are mostly detected at corner points, thus changing aspect ratio results in similar patches. Second, SIFT itself is a robust descriptor against moderate tilts, and low-rank SIFT even alleviates its sensitivty to local affine transforms. 

\section{Computing Low-Rank SIFT}
By way of analysis above, Low-rank SIFT is computed via the following three stages. First, feature points are detected using Harris Corner detector. Then, we perform low-rank transform on each feature point to locally rectify each patch. Finally, descriptors for the transformed point and patch are computed using SIFT. Note that the combination of Harris Corner Detector and SIFT is conventional which was introduced in~\cite{azad09}. \\\\
\textbf{The Computational Cost.}
To compute low-rank SIFT descriptors for an entire image of moderate size, as many as hundreds of low-rank optimizations on each of the feature point are expected. Following the Augmented Langrange Multiplier (ALM) solver proposed in~\cite{lin09}, each low-rank transform contains multiple iterations of singular value decomposition and is computationally intensive. Although local patch size is relatively small (in experiment we restrict the size to be $50\times 50$, as low-rank optimization would fail when the matrix size is overly small~\cite{rpca11}), it would still take up to 0.2 seconds per patch on average~\cite{harry12}. For an image with 500 feature points, it would take more than one minute to compute. This is infeasible for fast image retrieval. In light of this we developed methods to reduce the time cost.\\\\
\textbf{Low-Rank Integral Map.}
The low-rank integral map is a pre-processing step for fast computation of low-rank transform at any point, inspired by integral image widely used for fast computation of pixel sum of any region. We partition the image into $m\times n$ non-overlapping blocks, and for each block we compute a low-rank transform that rectifies that block. For any given point, we could find the block containing that point, and propagate the low-rank transform of the block to that point. The next theorem lays the foundation for such low-rank propagation. 

\textbf{Theorem 3.1. }\textit{Local low-rank transform at any point as its center could be approximated by the transform centered at a neighboring point.}

The theorem could be explained by the fact that, given any neighboring patch, we can assume they have almost identical affine property. That being said, $T_0$ which rectifies the patch around $p_0$ will rectify the patch around $p$ as well when applied. Therefore, $T$ could be approximated by $T_0$ plus a shift of center from $p_0$ to $p$, i.e. $T=T_0\circ T_t$, where $T_t$ represents the translation matrix from $p_0$ to $p$.

Empirically, if the image does not suffer from dramatic distortion, block size of $60\times 60$ usually works well for the purpose of approximating low-rank transform everywhere. In this way we reduce the number of low-rank optimizations to only a few.\\\\
\textbf{Fast Low-Rank Factorization.}
The computation of low-rank integral map could be even expedited, inspired by many of the recent fast low-rank factorization methods~\cite{naiyan12, zhou11}. One of the main ideas behind these methods is to pre-process the input matrix, either reduce its dimension or regularize its structure. We also observe that for adjacent patches, the affine property usually changes only slightly. Therefore we could use the already computed transform of neighboring block as prior information to pre-transform a block. Experiments show that it greatly reduces the iteration rounds as with this pre-processing step a block could already transformed to its rectified form. Another approach helpful to us is paralleled processing. This is possible as the optimization of blocks are mutually independent. In practice, we combine these two methods. We paralleling process each row and for each block in one row we process it sequentially and leverage computed transform of adjacent block. \\\\
The Low-rank SIFT algorithm is summarized in Algorithm~\ref{alg1}.
\begin{algorithm}[H]
\caption{Fast Low-rank SIFT Computation}
\label{alg1}
\begin{algorithmic}[1]
\STATE \textbf{Input:} Image $I$.\\
\STATE // Compute low-rank integral map $M$.
\STATE Partition $I$ into blocks $\{B_{ij}\}$ of size $60\times 60$.
\FOR{each row $i$}
\FOR{each block $j$ in row {i}}
\IF{j=1}
\STATE Compute low-rank transform $T_{ij}$ for $B_{ij}$.
\ELSE
\STATE Pre-transform $B_{ij}$ to $B_{ij}'$ using $T_{i(j-1)}$.
\STATE Compute low-rank transform $T_{ij}$ for $B_{ij}'$.
\ENDIF
\ENDFOR
\ENDFOR 
\STATE Low-rank integral map $M=\{T_{ij}\}$.
\STATE // Compute low-rank SIFT.
\STATE Detect interest points ${p_i}$ with Harris corner detector.
\FOR{each point $p_i$}
\STATE Let $P$ be the $50\times 50$ patch around $p_i$.
\STATE Let $T_i$ be the pre-computed low-rank transform, when $B_i$ is the block containing $p_i$. Let $T$ be the combined transform after translating $T_i$ to $p_i$. Compute $P'=T(P)$ and $p'=T(p_i)$.
\STATE Compute SIFT descriptor $v_i$ of point $p'$ and patch $P'$.
\ENDFOR
\STATE \textbf{Return:} Low-rank SIFT vector $V=\{v_i\}$.
\end{algorithmic}
\vspace{-0.2cm}
\end{algorithm} 

\section{Experiments}
We demonstrate the effectiveness of Low-rank SIFT by showing its superior performance in place recognition task. Buildings, unlike natural objects, are usually redundant in regularity, and are perfect candidate to utilize the power of Low-rank SIFT. This section is organized as follows. First, our method of constructing database of geotagged building images is described. Next, place recognition performance using Low-rank SIFT is compared against Harris corner, ASIFT and MSER detectors, all of which are combined with SIFT descriptor. Experiment results are also analyzed thereafter.\\\\
\textbf{Constructing Database of Geotagged Buildings.} 
The flourish of web geography services such as Google Earth and Bing Map makes it possible to fetch large geographical data from online directly. We propose a method to automatically construct database of geotagged building images by automatically downloading, extracting and processing images from Google Streetview. The database is then used in our place recognition experiment.

Given the coordinate of the city to construct the database with as input, our method consists of the following steps. 1. We sample 2D panorama images and 3D depth maps at the vicinity of the given location from Google Streetview. 3D depth maps provides depth information for each 2D image at pixel level. We will then get rectified building images using combined 2D-3D information. 2. Sky is segmented from 2D image using color. This is because 3D depth map is usually very inaccurate at top parts of buildings due to limited laser scan range, thus sky is used to refine the upper boundary. 3. We fit planes with 3D points and project the 3D planes back to 2D images. Fa\c{c}ades are cropped using both planes and sky. 4. Buildings of panoramic view are transformed to perspective view based on the geometry of panorama. 5. Perspective view is further rectified based on 3D points, whose spatial coordinates are used to compute the appropriate transform. Geolocation acquired with Google Streetview is saved along with the rectified image as well.

Table~\ref{database} listed the cities we have in our database and the quantity of rectified building images in each city. The quality of our image is not flawless. sometimes there are blurs and occlusions, and for some images the top part of buildings are rather obscure. Nevertheless we find it sufficient for effective place recognition. As the size of panorama image is very large, it also takes a significant amount of time to generate the database, where each city would take up to half a day to execute on an ordinary machine.
\begin{table}[ht]
\caption{Cities and their Image Quantity in Our Database}
\begin{center}
\resizebox{0.9\linewidth}{!}{
\begin{tabular}{|c|c||c|c|}
\hline
City &  Image Quantity &  City & Image Quantity\\
\hline
\hline
Paris& 1028& New York & 1387\\
\hline
Hong Kong	& 935 & Berlin & 821\\
\hline
London	& 1374 & Sydney & 1401\\
\hline
Pittsburg	& 1214 & Athens & 761\\
\hline
Seoul	& 1117 & Boston & 996\\
\hline
\end{tabular}
}
\end{center}
\label{database}
\vspace{-0.4cm}
\end{table}\\\\
\textbf{Place Recognition with Feature Selection.}
We test place recognition using Low-rank SIFT and other features using acquired database. Vocabulary tree model for object recognition~\cite{nister06a} is employed as it is scalable and fast enough for database of large size, as is in our case. The vocabulary tree is constructed using features of images in database and feature vector of query image is supplied and traverses the tree, and the nodes visited are used to compare and compute their similarity scores.

We adopt a novel scheme to select useful features from query image, and found it capable of improving our recognition performance significantly (Table~\ref{perf}). The reason for feature selection is that query images, unlike those in database which are mostly cropped and clean, usually suffer from noise and occlusions (sky, tree and pedestrian). Inspired by the low-rank property of local patch of buildings, we only keep interest points of those with patches of certain rank, and filter the rest. Typically valid rank should not be too small which corresponds to homogeneous color (sky), nor should it be overly large which corresponds to natural objects. In experiment we found retaining patches of rank between 2 to 5 generally works well for feature selection purpose. 

\begin{figure}[t]
\begin{center}
\includegraphics[width=0.9\linewidth]{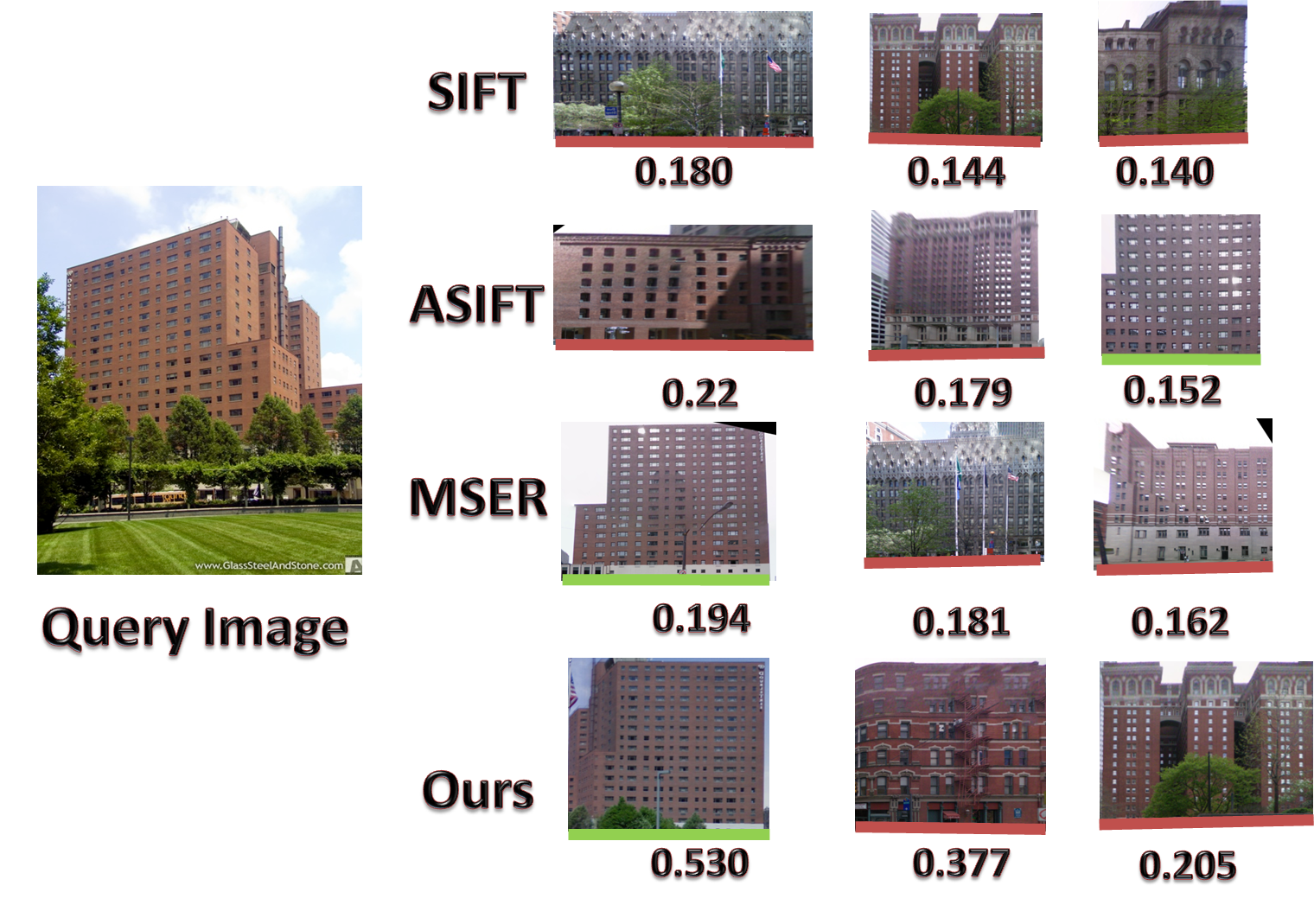}
\caption{Place recognition results returned using different features. Top three closest matches of each feature with their similarity scores are shown.}
\label{comparison}
\end{center}
\vspace{-0.4cm}
\end{figure}

Performance of different features are listed and compared in Table~\ref{perf} and Fig.~\ref{comparison}. Our query images are randomly selected from online, containing both buildings and other scenes. We see significant performance improvement by using Low-rank SIFT. Our result also beats previous place recognition performance benchmark of~\cite{janknopp10}. As expected, improvements are mostly contributed by building queries, i.e., those with regular structures and able to leverage Low-rank SIFT, rather than queries of open scenes. An example is shown in Fig.~\ref{comparison}, where perspective view of DoubleTree hotel is given as input, and only Low-rank SIFT successfully identifies the location. 
\begin{table}[ht]
\caption{The percentage of correctly localized queries by using various features(measured in \%).}
\begin{center}
\resizebox{0.9\linewidth}{!}{
\begin{tabular}{|c|c|c|c|}
\hline
\textbf{Feature} &  Overall & Building Query &  Scene Query\\
\hline
\hline
Harris Corner	& {20.83}	& 29.74 & 10.35 \\
\hline
ASIFT	& {29.78}  & 38.56 & 20.81\\
\hline
MSER	& {26.14} & 31.72 & 21.24 \\
\hline
Low-rank SIFT	& \textbf{35.69/40.25(with feature selection)} & \textbf{60.93} & \textbf{20.21}\\
\hline
\end{tabular}
}
\\
\end{center}
\label{perf}
\vspace{-0.2cm}
\end{table}

The success of Low-rank SIFT comes at no surprise. By carefully examining the distance between query image and database, we found the minimum distance, i.e. distance of the closest match, drops as much as 50\% after applying Low-rank SIFT. More concrete analysis indicates that once rectified, most query features would fall in the same node with their correctly matched features in database. Our method is also efficient, as computation of Low-rank SIFT plus image retrieval could mostly be accomplished within 5 seconds. Failure cases are usually caused by severe image distortion that could not be recovered using local affine transform, or when the scene is undermined by large occlusions.
\section{Conclusion}
In this paper, we propose a novel affine invariant feature for place recognition by exploiting the local low-rank property of man-made objects. The feature could be computed efficiently using our Low-rank Integral Map approach and be further compressed based on the rank of local patches. Experiment shows its superior performance against other conventional features. As future work, we plan to develop the feature to expand its usage to other computer vision tasks, such as image matching and repetition detection.

\bibliographystyle{IEEEtran}      

\bibliography{egbib}

\end{document}